\documentclass[10pt,twocolumn,letterpaper]{article}

\usepackage{cvpr}
\usepackage{times}
\usepackage{epsfig}
\usepackage{graphicx}
\usepackage{amsmath}
\usepackage{amssymb}
\usepackage{afterpage}
\usepackage{placeins}
\usepackage{pgfplots,pgfplotstable,authblk}

\usepackage[pagebackref=true,breaklinks=true,letterpaper=true,colorlinks,bookmarks=false]{hyperref}

 \cvprfinalcopy 

\def\cvprPaperID{309} 

\newcommand{\boldhead}[1]{\vspace{0.02in}\noindent\textbf{#1}:}
\newcommand{\boldheadr}[1]{\vspace{0.02in}\noindent\textbf{#1}}

\ifcvprfinal\fi
\begin{document}

\title{Where To Look: Focus Regions for Visual Question Answering}

\author{Kevin J. Shih}
\author{Saurabh Singh}
\author{Derek Hoiem}
\affil{University of Illinois at Urbana-Champaign}
\affil{\tt\small \{kjshih2, ss1, dhoiem\}@illinois.edu}


\maketitle

\begin{abstract}
We present a method that learns to answer visual questions by
selecting image regions relevant to the text-based query. Our method
maps textual queries and visual features from various regions into a shared space where they are
compared for relevance with an inner product. Our method
exhibits significant improvements in answering questions such as ``what
color,'' where it is necessary to evaluate a specific location,
and ``what room,'' where it selectively identifies informative image
regions. Our model is tested on the recently released VQA~\cite{VQA}
dataset, which features free-form human-annotated questions and
answers. 
\end{abstract}

\section{Introduction}

\begin{figure}
  \centering
  \includegraphics[width=\columnwidth]{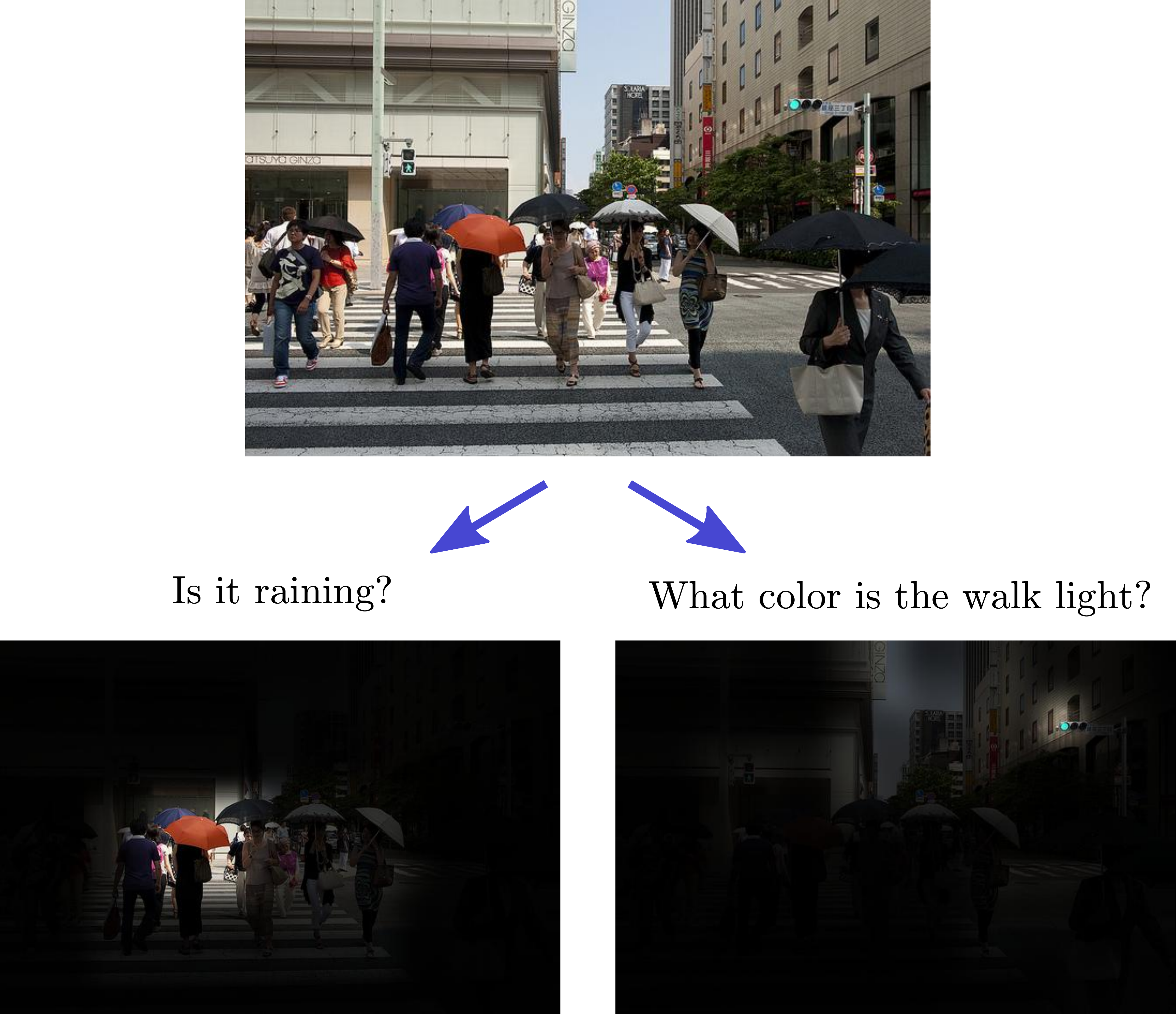}
  \caption{Our goal is to identify the correct answer for a natural
    language question, such as ``What color is the walk light?'' or
    ``Is it raining?'' We particularly focus on the problem of
    learning where to look.  This is a challenging problem as it
    requires grounding language with vision and learning to recognize
    objects, use relations, and determine relevance. For example,
    whether it is raining may be determined by detecting the presence of puddles
    gray skies, or umbrellas in the scene, whereas the color of the walk light
    requires focused attention on the light alone. The above figure
    shows example attention regions produced by our proposed model.}
  \label{fig:rain_scene}
\end{figure}

Visual question answering (VQA) is the task of answering a natural language
question about an image. VQA includes many challenges
in language representation and grounding, recognition, common sense
reasoning, and specialized tasks like counting and reading.  In this
paper, we focus on a key problem for VQA and other visual reasoning
tasks: knowing where to look. Consider Figure~\ref{fig:rain_scene}. It's
easy to answer ``What color is the walk light?'' if the light bulb is localized,
while answering whether it's raining may be dealt with by identifying
umbrellas, puddles, or cloudy skies. We want to learn where to look to
answer questions supervised by only images and question/answer pairs.
For example, if we have several training examples for ``What time of
day is it?'' or similar questions, the system should learn what kind
of answer is expected and where in the image it should base its
response.


\begin{figure}[t!]
  \centering
  \includegraphics[width=\linewidth]{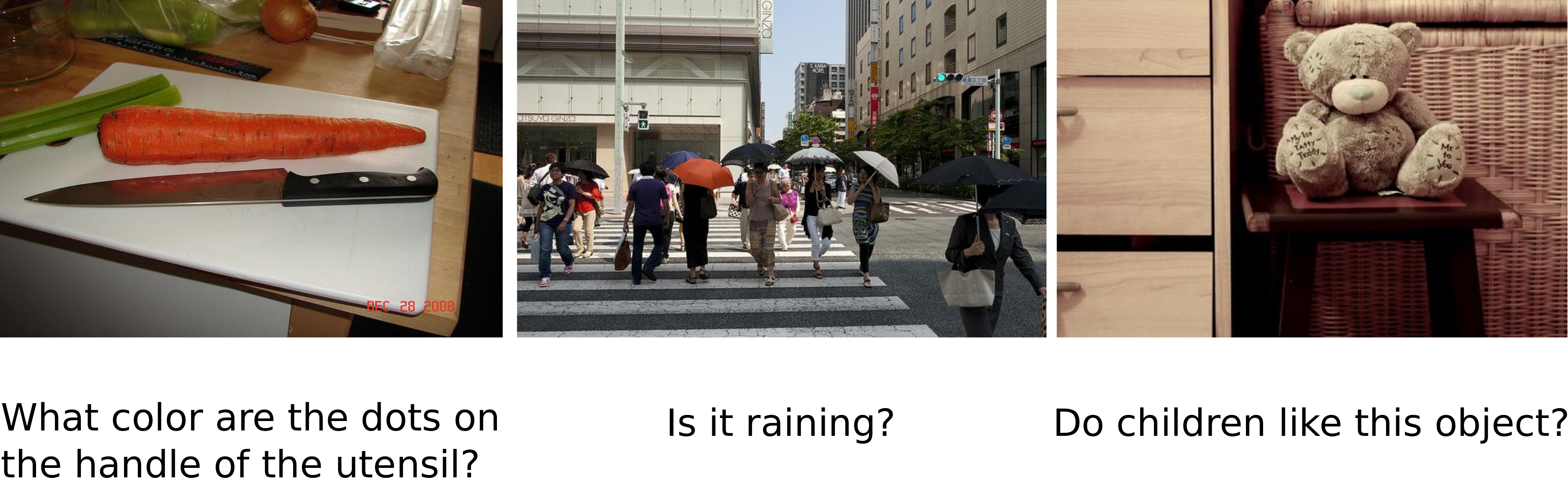}
  \caption{Examples from VQA~\cite{VQA}. From left to right, the above examples require focused region information to pinpoint the dots, whole image information to determine the weather, and abstract knowledge regarding relationships between children and stuffed animals.}
  \label{fig:example}
\end{figure}

Learning where to look from question-image pairs has many
challenges. Questions such as ``What sport is this?'' might be
best answered using the full image.  Other questions such as ``What
is on the sofa?'' or ``What color is the woman's shirt?'' require
focusing on particular regions.  Still others such as ``What does the
sign say?'' or ``Are the man and woman dating?'' require specialized
knowledge or reasoning that we do not expect to achieve. The system
needs to learn to recognize objects, infer spatial
relations, determine relevance, and find correspondence between
natural language and visual features.  Our key idea is to learn a
non-linear mapping of language and visual region features into a
common latent space to determine relevance. The relevant regions are
then used to score a specific question-answer pairing. The latent embedding and
the scoring function are learned jointly using a margin-based loss
supervised solely by question-answer pairings. We perform
experiments on the VQA dataset~\cite{VQA} because it features
open-ended language, with a wide variety of questions (see
Fig.~\ref{fig:example}).  Specifically, we focus on its
multiple-choice format because its evaluation is much less ambiguous
than open-ended answer verification.

We focus on learning where to look but also provide useful baselines and analysis for the task as a whole.  Our contributions are as follows:
\begin{itemize}
\item We present an image-region selection mechanism that learns
  to identify image regions relevant to questions.
\item We present a learning framework for solving multiple-choice
  visual QA with a margin-based loss that significantly outperforms
  provided baselines from~\cite{VQA}.
\item We compare with baselines that answer questions without the
  image, use the whole image, and use all image regions with uniform
  weighting, providing a detailed analysis for when selective regions improve VQA performance.
\end{itemize}

\section{Related Works}
Many recent works in tying text to images have
explored the task of automated image captioning \cite{gong2014improving, fang2014captions,  xu2015show,
  karpathydeep, mao2014deep, mao2014explain, lrcn2014, Chen_2015_CVPR,
Vinyals_2015_CVPR}. While VQA can be considered as a type of directed
captioning task, our work relates to some~\cite{xu2015show, fang2014captions} in that we learn to employ an attention mechanism
for region focus, though our formulation makes determining region
relevance a more explicit part of the learning process. In Fang et al.~\cite{fang2014captions}, words
are detected in various portions of the image and combined together
with a language model to generate captions. Similarly, Xu et al.~\cite{xu2015show}
uses a recurrent network model to detect salient objects and generate
caption words one by one.  Our model works in
the opposite direction of these caption models at test time by determining the relevant image
region given a textual query as input. This allows our model to
determine whether a question-answer pair is a good match given evidence from
the image.

Partly due to the difficulty of evaluating image captioning, several
visual question answering datasets have been proposed along with
applied approaches.  We choose to experiment on VQA~\cite{VQA} due to
the open ended nature of its question and answer
annotations. Questions are collected by asking annotators to pose a difficult question for a smart robot, and multiple
answers are collected for each question.  We experiment on the
multiple-choice setting as its evaluation is less ambiguous than that
of open-ended response evaluation.  Most other visual question answering
datasets~\cite{ren2015image,yu2015visual} are based on reformulating
existing object annotations into questions, which provides an
interesting visual task but limits the scope of visual and abstract
knowledge required.

Our model is inspired by End-to-End Memory Networks~\cite{SukhbaatarSWF15} proposed
for answering questions based on a series of sentences.  The regions
in our model are analogous to the sentences in theirs, and, similarly
to them, we learn an embedding to project question and potential
features into a shared subspace to determine relevance with an inner product.  Our method
differs in many details such as the language model and more broadly in
that we are answering questions based on an image, rather than a text
document. Ba et al.~\cite{ba2015predicting} also uses a similar architecture, but in a zero-shot
learning framework to predict classifiers for novel categories. They project language and vision features into a shared subspace to perform similarity computations with inner products like us, though the score is used to guide the generation of object classifiers rather than to rank image regions.

Existing approaches in VQA tend to use recurrent
networks to model language and predict
answers~\cite{ren2015image,VQA,yu2015visual,malinowski2015iccv}, though simpler Bag-Of-Words (BOW) and averaging models have been
shown to perform roughly as well if not better than sequence-based
LSTM~\cite{ren2015image, VQA}. Yu et al.~\cite{yu2015visual}, which 
proposes a Visual Madlibs dataset for fill-in-the-blank and question
answering, focuses their approach on learning latent embeddings and 
finds normalized CCA on averaged word2vec representations~\cite{gong2014improving,mikolov2013efficient}  to outperform recurrent networks for embedding. Similarly in our work, we find a fixed-length averaged representation of word2vec vectors for language to be
highly effective and much simpler to train, and our approach differs at a high level in our
focus on learning where to look.

\section{Approach}
\begin{figure*}[ht]
  \centering
  \includegraphics[width=\linewidth]{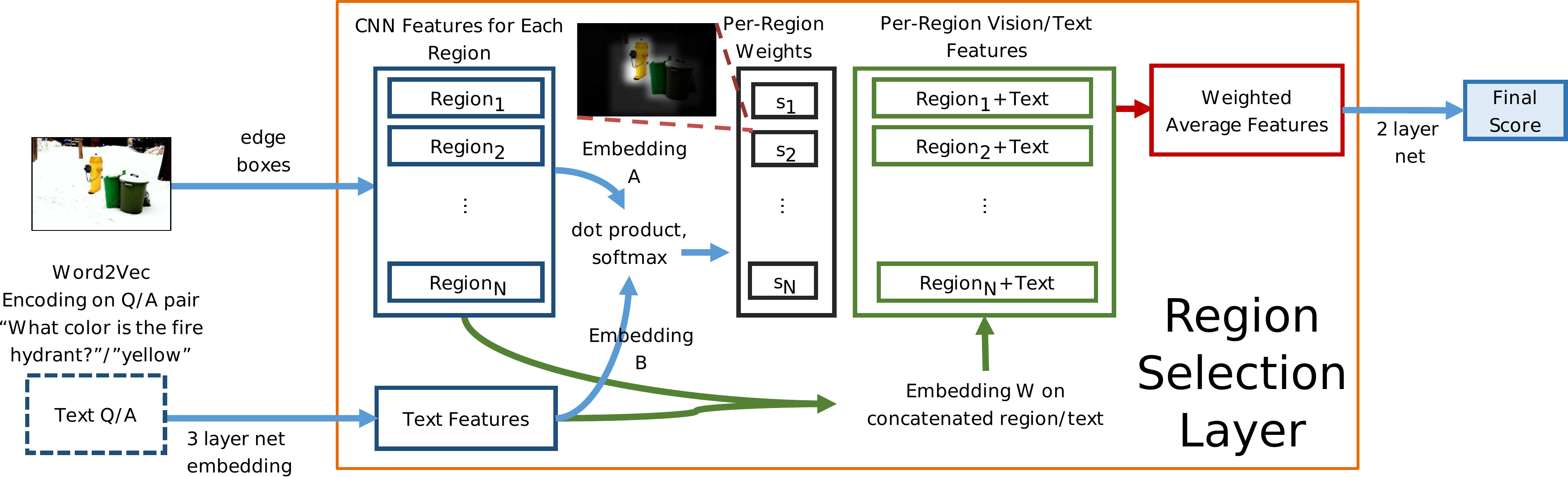}
  \caption{Overview of our network for the
    example question-answer pairing: ``What color is the fire hydrant? Yellow.'' Question
    and answer representations are concatenated, fed through the network, then
    combined with selectively weighted image region features to
    produce a score.}
  \label{fig:overview}
\end{figure*}

Our method learns to embed the textual question and the
set of visual image regions into a latent space where the inner product
yields a relevance weighting for each region. See
Figure~\ref{fig:overview} for an overview. The input is a question,
potential answer, and image features from a set of automatically selected candidate regions.  We encode the parsed question and answer using word2vec~\cite{mikolov2013efficient} and a two-layer network.  Visual features for each region are encoded using the top two layers (including the output layer) of a CNN trained on ImageNet~\cite{ILSVRC15}.  The language and vision features are then embedded and compared with a dot product, which is soft-maxed to produce a per-region relevance weighting.  Using these weights, a weighted average of concatenated vision and language features is the input to a 2-layer network that outputs a score for whether the answer is correct.

\subsection{QA Objective}
Our model is trained for the multiple choice task of the VQA
dateset. For a given question and its corresponding choices, the
objective of our network aims to maximize a margin between correct and
incorrect choices in a structured-learning fashion.  We achieve this
by using a hinge loss over predicted confidences $y$.

In our setting, multiple answers could be acceptable to varying
degrees, as correctness is determined by the consensus of 10
annotators.  For example, most may say that the color of a scarf is
``blue'' while a few others say ``purple''.  To take this into
account, we scale the margin by the gap in number of annotators returning the specific answer:
\begin{equation}
  {\mathcal L}(y) = \max_{\forall n \ne p} (0,y_n + (a_p - a_n) - y_p).
\end{equation}

The above objective requires that the score of the correct answer ($y_p$) is at
least some margin above the score of the highest-scoring incorrect answer
($y_n$) selected from among the set of incorrect choices ($n \ne p$).
For example, if 6/10 of the annotators answer $p$ ($a_p = 0.6$) and
2 annotators answer $n$ ($a_n = 0.2$), then $y_p$ should outscore $y_n$ by a margin of at least 0.4.

\subsection{Region Selection Layer}

Our region selection layer selectively combines
incoming text features with image features from relevant regions of
the image.  To determine relevance, the layer first projects the image
features and the text features into a shared
N-dimensional space, after which an inner product is computed for each
question-answer pair and all available regions.

Let $G_r$ be the projection of all region features in column vectors
of $X_r$, and $\vec{g}_l$ be the projection of a single embedded question-answer
pair. The feedforward pass to compute the relevance weightings is
computed as follows:
\begin{align}
  G_r =& AX_r+\vec{b}_r\\
  \vec{g}_l =& B\vec{x}_l + \vec{b}_l\\
  \vec{s}_{l,r} =& \sigma(G_r^T\vec{g}_l)\\
  \sigma(\vec{z}) = &\frac{e^{z_j}}{\sum_{k=1}^{K}e^{z_k}}\mbox{ for }
  j = 1,...K
\end{align}
Here, the output $\vec{s}_{l,r}$ is the softmax normalized weighting
($\sigma$) of the inner products of $\vec{g}_l$ with each projected
region feature in $G_r$. Vectors $\vec{b}$ represent biases. The
purpose of the inner product is to force the model to determine region
relevance in a vector similarity fashion. 

Using 100 regions per image, this gives us 100 region weights for a question-answer pair. Next, the text features are concatenated directly with image features for each region to produce
100 different feature vectors.  This is shown in the horizontal
stacking of $X_r$ and repetitions of $\vec{x}_l$ below. Each feature vector is linearly projected with $W$, and the weighted average is computed using $\vec{s}_r$ to attain feature vector $\vec{a}_l$ for each question and answer pair, which is then fed through relu and batch-normalization layers. 

\begin{align}
  P_{l,r} =& W\left[ {\begin{array}{c} X_r\\\begin{array}{ccc}- &\vec{x}_l &-\end{array} \end{array}} \right]+\vec{b}_o\\
  \vec{a}_l =& P\vec{s}_{l,r}
\end{align}

We also tried learning to predict a relevance score directly from concatenated vision and language features, rather than computing the dot product of the features in a latent embedded space.  However,
the resulting model appeared to learn a salient region weighting
scheme that varied little with the language component.  The
inner-product based relevance was the only formulation we tried that successfully takes account of both the query and the region information.

\subsection{Language Representation}

We represent our words with 300-dimensional word2vec
vectors~\cite{mikolov2013efficient} for their simplicity 
and compact representation.  We are also motivated by the ability of vector-based language representations to encode similar words with similar vectors, which may aid answering open-ended questions. 
Using averages across word2vec vectors, we construct fixed-length
vectors for each question-answer pair, which our model then learns to
score. In our results section, we show that our vector-averaging language
model noticeably outperforms a more complex LSTM-based model from
~\cite{VQA}, demonstrating that BOW-like models provide very effective
and simple language representations for VQA tasks.

We first tried separately averaging vectors for each word with the
question and answer, concatenating them to yield a 600-dimensional
vector, but since the word2vec representation is not sparse, averaging
several words may muddle the representation.  We improve the
representation using the Stanford Parser~\cite{de2006generating} to bin the question into
additional separate semantic bins. The bins are defined as follows:

\boldheadr{Bin 1} captures the type of question by averaging
  the word2vec representation of the first two words.  For example, ``How many'' tends to require a numerical answer,
  while ``Is there'' requires a yes or no answer.\\
\boldheadr{Bin 2} contains the nominal subject to encode subject of question.\\
\boldheadr{Bin 3} contains the average of all other noun words.\\
\boldheadr{Bin 4} contains the average of all remaining words,
excluding determiners such as ``a,'' ``the,'' and ``few.''

\begin{figure}[ht]
  \centering
  \includegraphics[width=0.7\columnwidth]{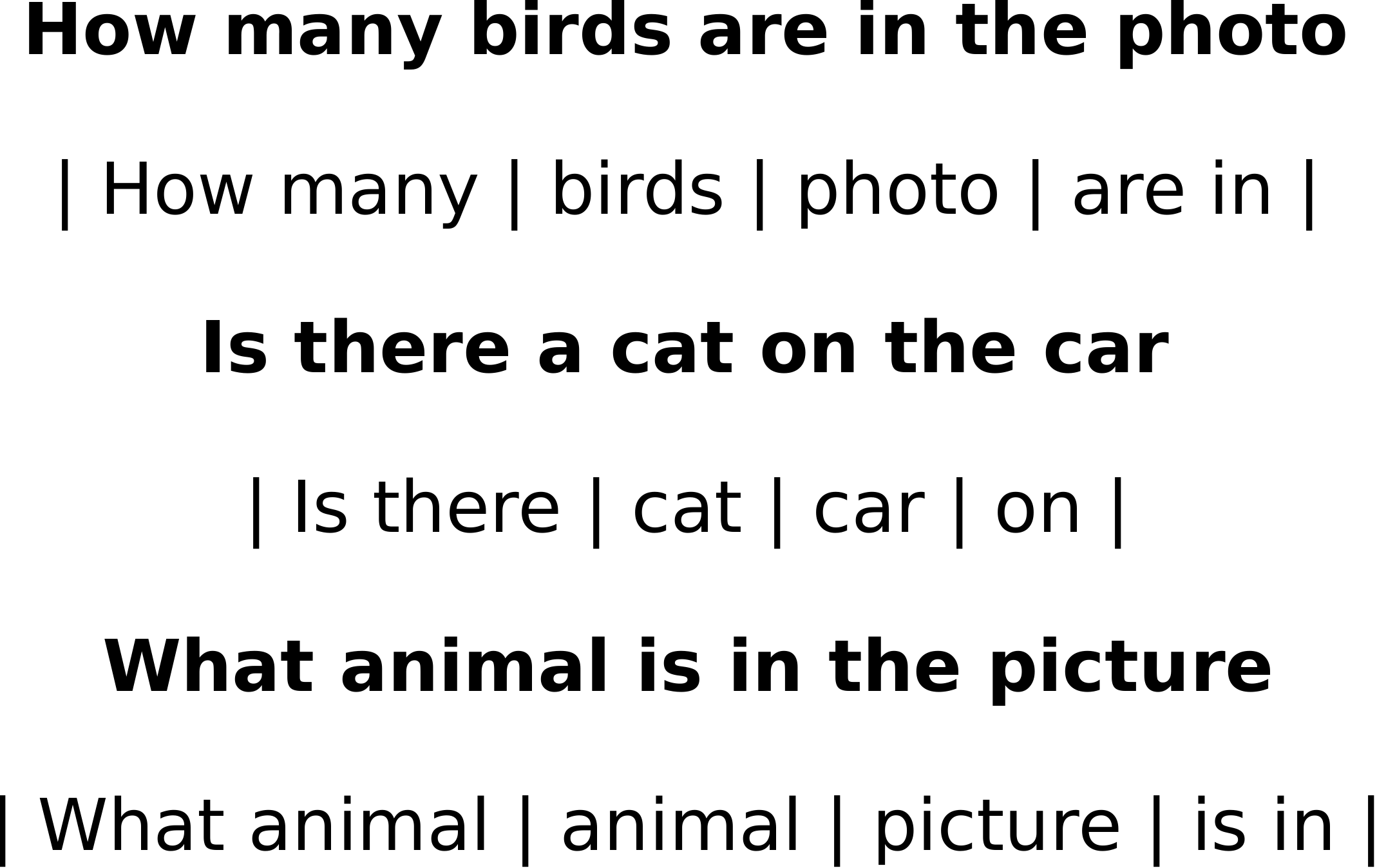}
  \caption{Example parse-based binning of questions. Each bin is represented with the average of the word2vec vectors of its members. Empty bins are represented with a zero-vector.}
  \label{fig:parsing_example}
\end{figure}

Each bin then contains a 300-dimensional representation, which are
concatenated with a bin for the words in the candidate answer to yield
a 1500-dimensional question/answer representation.  Figure~\ref{fig:parsing_example} shows examples of binning for the parsed question.  This representation separates out important components of a variable-length question while maintaining a fixed-length representation that simplifies the network architecture.

\subsection{Image Features}

The image features are fed directly into the region-selection
layer from a pre-trained network.  We first select candidate regions
by extracting the top-ranked 99 Edge Boxes~\cite{zitnick2014edge} from
the image after performing non-max suppression with a 0.2 intersection
over union overlap criterion. We found this aggressive non-max
suppression to be important for selecting smaller regions that may be
important for some questions, as the top-ranked regions tend to be
highly overlapping large regions. Finally, a whole-image region is also added to
ensure that the model at least has the spatial support of the full
frame if necessary, bringing the total number of candidate regions to
100 per image. While we have not experimented with the number of
regions, it is possible that the improved recall from additional regions
may improve performance.

We extract features using the VGG-s network~\cite{Chatfield14}, concatenating the output
from the last fully connected layer (4096 dimensions) and the
pre-softmax layer (1000 dimensions) to get a 5096 dimensional feature
per region.  The pre-softmax classification layer was included to
provide a more direct signal for objects from the Imagenet~\cite{ILSVRC15} classification task.

\subsection{Training}

Our overall network architecture is multi-layer feed-forward network
as seen in Fig.~\ref{fig:overview}, implemented in
MatConvNet\cite{vedaldi15matconvnet}. Our fully connected layers are
initialized with Xavier initialization ($\frac{1}{\sqrt{n_{in}}}$)~\cite{glorot2010understanding} and
separated with a batch-normalization~\cite{ioffe2015batch} and relu
  layer~\cite{glorot2011deep} between each. The word2vec text features
  are fed into the network's input layer, whereas the image region features
feed in through the region selection layer.

Our network sizes are set as follows. The 1500 dimensional language
features first pass through 3 fully connected layers with output
dimensions 2048, 1500, and 1024 respectively. The embedded language features are then passed through the region selection layer to be combined with the vision
features. Inside the region selection layer, projections $A$ and $B$
project both vision and language representations down to 900
dimensions before computing their inner product. The exiting feature
representation passes through $W$ with an output dimension of
2048. then finally through two more fully connected layers with output
dimensions of 900 and 1 where the output scalar is the
question-answer pair score.

It is necessary to pay extra attention to the initialization of the
region-selection layer. The magnitude of the projection matrices $A$,
$B$ and $W$ are initialized to $0.001$ times the standard normal
distribution.  We found that low initial values were important to
prevent the softmax in selection from spiking too early and to prevent
the higher-dimensional vision component from dominating early in the training.


\section{Experiments}
We evaluate the effects of our region-selection layer on the
multiple-choice format of the MS COCO Visual Question Answering (VQA)
dataset~\cite{VQA}. This dataset contains 82,783 images for training,
40,504 for validation, and 81,434 for testing.  Each image has 3
corresponding questions with recorded free-response answers from
10 annotators. Any response that comes from at least 3 annotators is
considered correct. We use the 18-way multiple choice task because its
evaluation is much less ambiguous than the open-ended response task,
though our method could be applied to the latter by treating the most
common or likely K responses as a large K-way multiple
choice task.  We trained using only the training set, with 10\% set
aside for model selection and parameter tuning.  We perform detailed
evaluation on the validation set and further comparison on the test
set using the provided submission tools.

\begin{figure}
  \centering
  \includegraphics[width=\columnwidth]{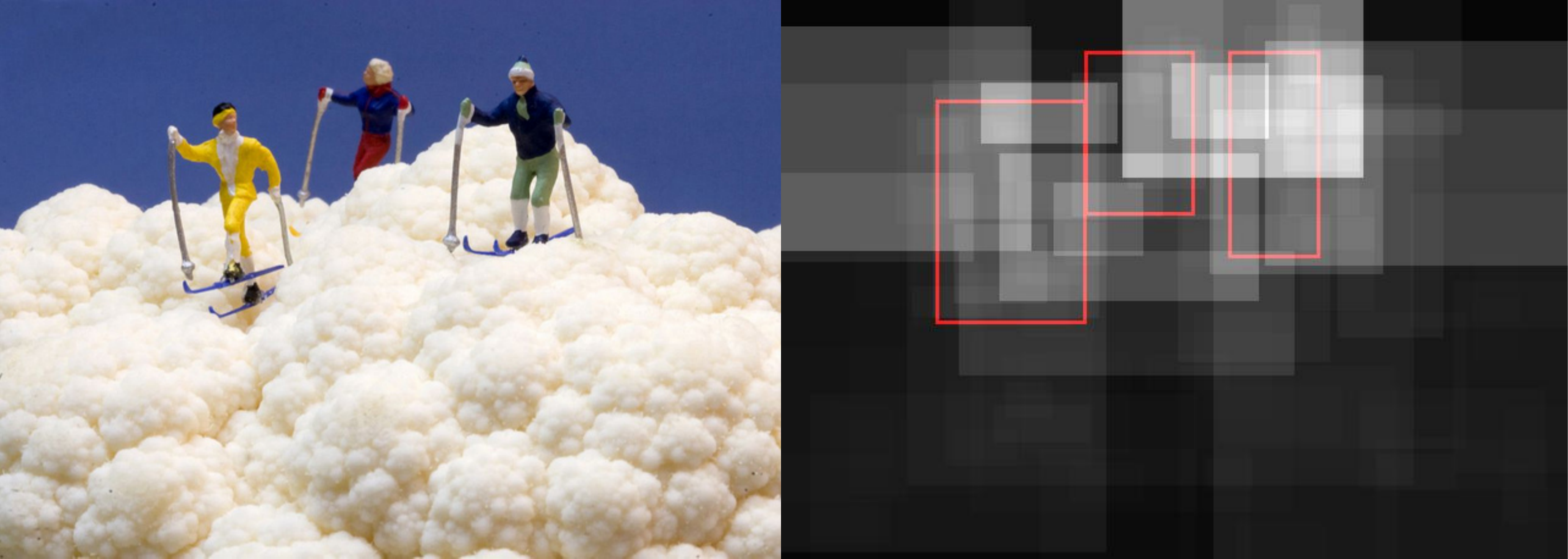}
  \caption{Example image with corresponding region weighting. Red
    boxes correspond to manual annotation of regions relevant to the
    question: ``Are the people real?''}
  \label{fig:weight_anno}
\end{figure}

We evaluate and analyze how much our region-weighting improves accuracy  compared to using the whole image or only language (Tables~\ref{tbl:Overall_acc_val},~\ref{tbl:Overall_acc_testdev},~\ref{tbl:breakdown}) and show examples in Figure~\ref{fig:vqa_qual_comp}.  We also perform a simple evaluation on a subset of images showing that relevant regions tend to have higher than average weights (Fig.~\ref{fig:weight_anno}).  We also show the advantage of our language model over other schemes (Table~\ref{tbl:language}).  

\subsection{Comparisons between region, image, and language-only models}

We compare our region selection model with several baseline methods, described below.

\boldhead{Language-only} We train a network to score each answer purely from the language representation.  This provides a baseline to demonstrate improvement due to image features, rather than just good guesses.

\boldhead{Word+Whole image} We concatenate CNN features computed over the entire image with the language features and score them using a 3-layer neural network, essentially replacing the region-selection layer with features computed over the whole image.  

\boldhead{Word+Uniform averaged region features} To test that region weighting is important, we also try uniformly averaging features across all regions as the image representation and train as above.

\begin{table}
  \centering
\begin{tabular}{l|c}
  Model & Overall (\%)\\
  \hline
  Language Only & 53.98\\
  Word+Whole Image & 57.83\\
  Word+ave. reg. & 57.88\\
  Word+Region Sel. & \textbf{58.94}\\
  \hline
  LSTM Q+I~\cite{VQA} & 53.96\\
\end{tabular}
\caption{Overall accuracy comparison on Validation. Our region
  selection model outperforms our own baselines, demonstrating the
  benefits of selective region weighting.}
\label{tbl:Overall_acc_val}
\end{table}

Table~\ref{tbl:Overall_acc_val} shows the comparison of overall accuracy
on the validation set.  Our proposed region-selection model outperforms all other models.  Also, we can see that uniform weighting of regions is not helpful.  We also include the
best-performing LSTM question+image model from the authors of the VQA
dataset~\cite{VQA}. This model significantly underperforms even our much simpler baselines, which could be partly because the model was designed for open-ended answering and adapted for multiple choice.

We also evaluate our model on the test-dev and test-standard partitions in order to compare with
additional models from~\cite{VQA}. In Table~\ref{tbl:Overall_acc_testdev},
we include comparisons to the best-performing question+image based
models from the VQA dataset paper~\cite{VQA}. Our model was retrained on train+val data, using a 10\% held-out set from the train set for model selection. Note that our model significantly outperforms the baselines in the ``others'' category, which contains the majority of the question types that our model excels at.

\begin{figure*}[ht!]
  \centering
  \includegraphics[width=0.86\linewidth]{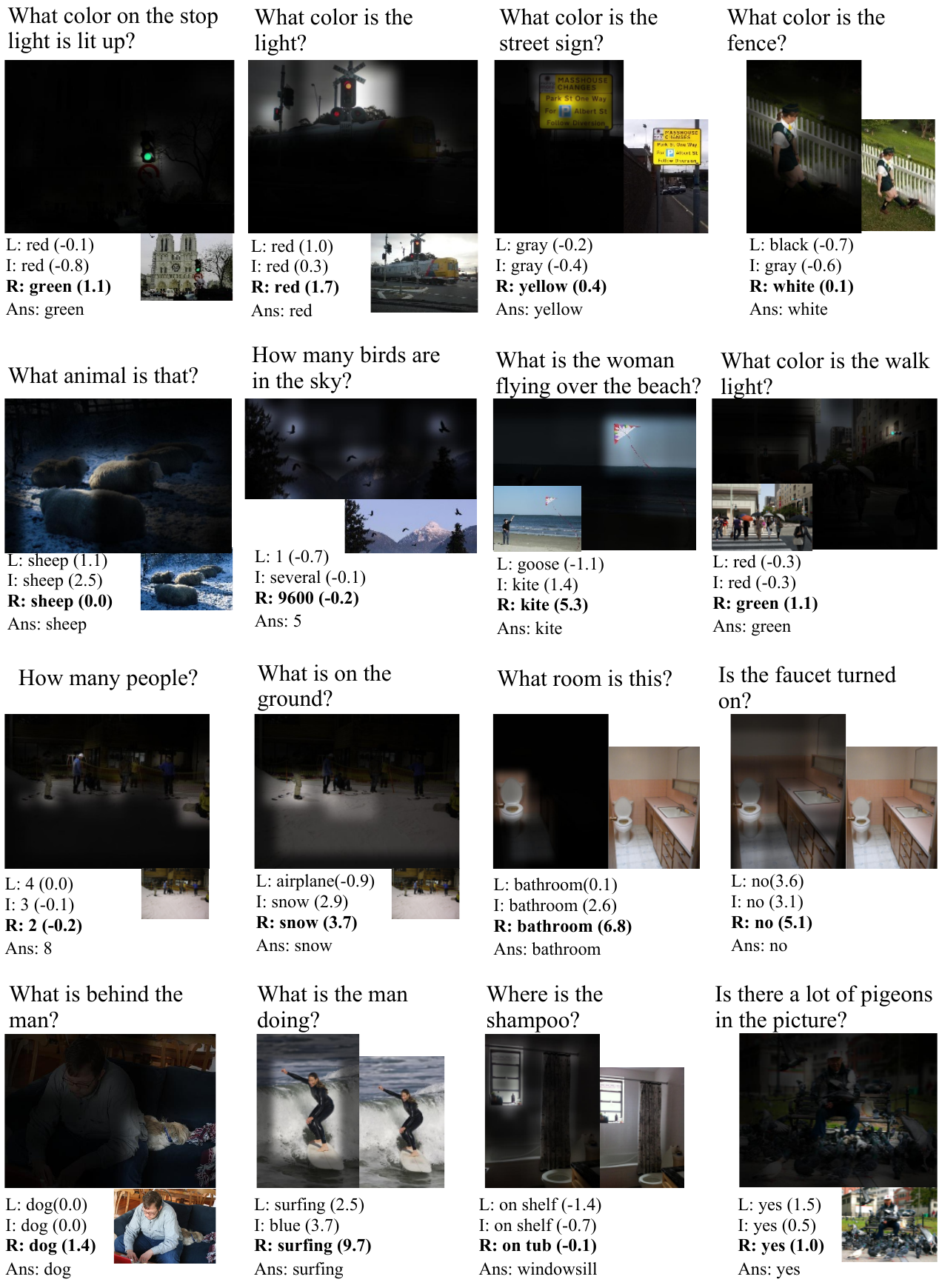}
  \caption{Comparison of qualitative results from Val. The larger image shows the
    selection weights overlayed on the original image (smaller). L:
    Word only model; I: Word+Whole Image; R: Region Selection. The
    scores shown are ground truth confidence - top incorrect. Note
    that the first row shows successful examples in which tight region
  localization allowed for an accurate color detection. In the third
  row, we show examples of how weighting varies on the same image due
  to differing language components. }
  \label{fig:vqa_qual_comp}
\end{figure*}

Table~\ref{tbl:breakdown} offers a more detailed performance summary
across various question types, with discussion in the caption.
Figure~\ref{fig:vqa_qual_comp} shows a qualitative comparison of
results, highlighting some of the strengths and remaining problems of
our approach. These visualizations are created by soft masking the
image by with a mask created by summing the weights of each region and
normalizing to a max of one.  A small blurring filter is applied to
the soft mask before normalization to remove distracting artifacts that occur from multiple overlapping rectangles.
On color questions, localization of the mentioned object tends to be very good, which leads to much more accurate answering.  On some questions, such as ``How many birds are in the sky?'' the system cannot produce the correct answer but does focus on the relevant objects.  The third row shows examples of how different questions lead to different focus regions.  Notice how the model identifies the room as a bathroom in the third row by focusing on the toilet, and, when confirming that ``kite'' is the answer to ``What is the woman flying over the beach?'' focuses on the kite, not the woman or the beach.

\begin{figure*}
  \includegraphics[width=0.95\linewidth]{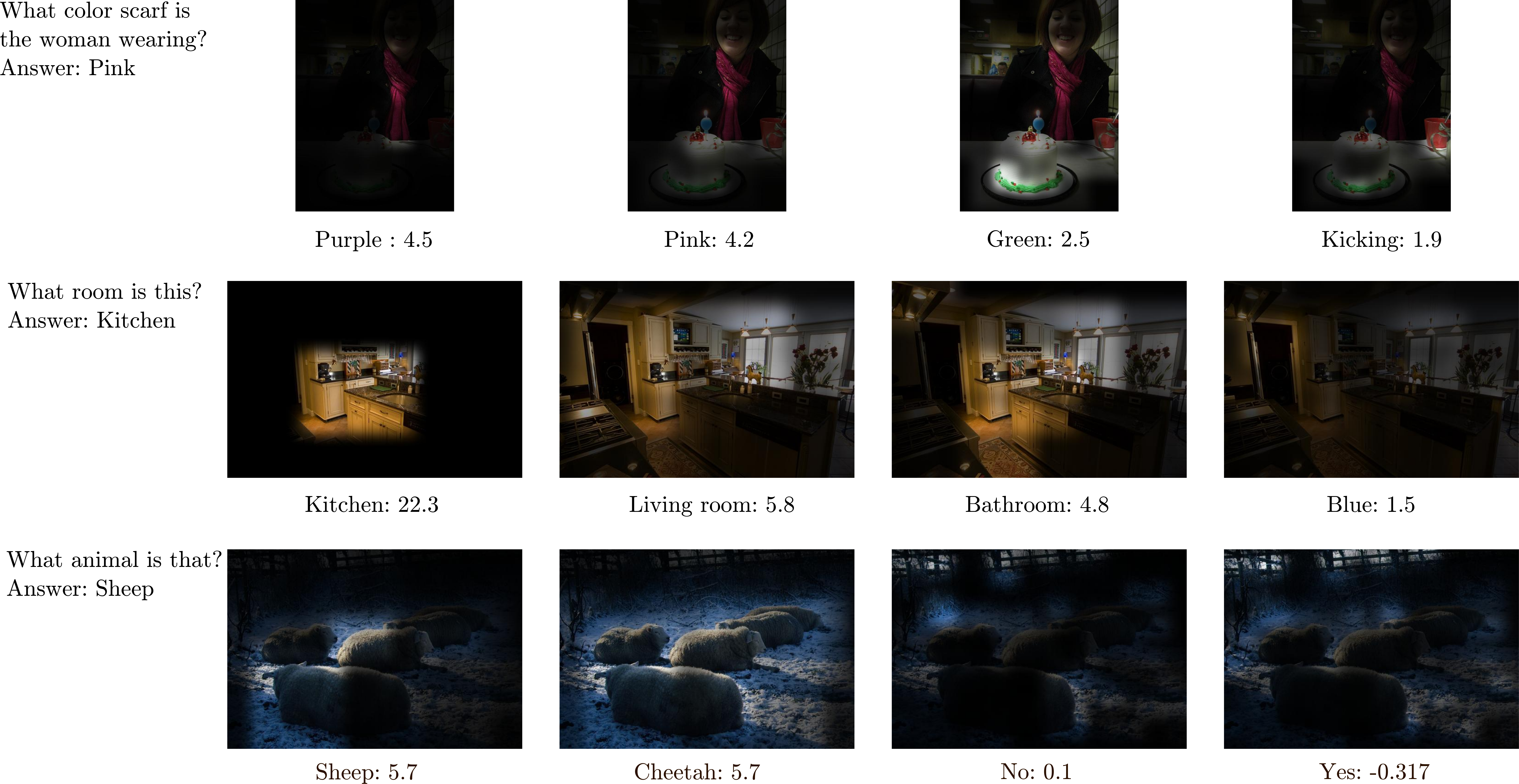}
  \caption{Comparison of attention regions generated by various
    question-answer pairings for the same question. Each visualization
    is labeled with its corresponding answer choice and returned
    confidence. We show the highlighted regions for the top multiple
    choice answers and some unrelated ones. Notice that in the first
    example, while the model clearly identified a green region within
    the image to match the ``green'' option, the corresponding
    confidence was significantly lower than that of the correct
    options, showing that the model does more than just match answer
    choices with image regions.}
  \label{fig:mc_diff}
\end{figure*}

\begin{table}
  \centering
\begin{tabular}{l|c|c|c|c}
  Model & All & Y/N & Num. & Others\\
  \hline
  Word+Region Sel. & \textbf{62.44} & 77.62 & 34.28 & \textbf{55.84}\\
  LSTM Q+I~\cite{VQA} & 57.17 & \textbf{78.95} & \textbf{35.80} & 43.41\\
  Q+I~\cite{VQA} & 58.97 & 75.97 & 34.35 & 50.33\\
  \hline
  Word+Region Sel. & 62.43 & 77.18 & 33.52 & 56.09\\
\end{tabular}
\caption{Accuracy comparison on Test-dev (top) and Test-standard (bottom). Our model
  outperforms the best performing image and text models from ~\cite{VQA}.}
\label{tbl:Overall_acc_testdev}
\end{table}

In Figure~\ref{fig:mc_diff}, we show additional qualitative examples
of how the region selection varies with question-answer pairs. In the
first row, we see the model does more than simply match answer choices
to regions. While it does find a matching green region, the
corresponding confidence is still low. In addition, we see that
irrelevant answer choices tend to have less-focused attention
weightings. For example, the kitchen recognition question has most of
its weighting on what appears to be a discriminative kitchen patch for
the correct choice, whereas the ``blue'' choice appears to have a more
evenly spread out weighting.

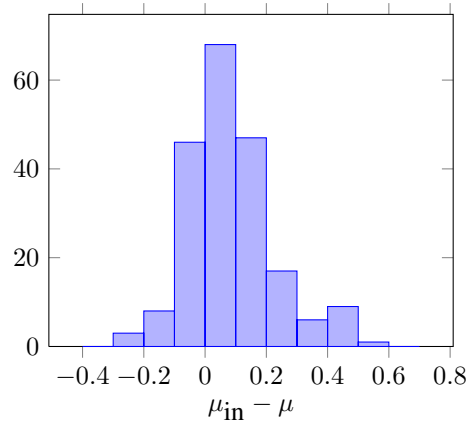
\begin{figure}[ht]
  \centering
\begin{tikzpicture}
  \begin{axis}[
      ybar,
      ymin=0,
      xlabel=$\mu_{\mbox{in}} - \mu$,
      height=6cm
    ]
    \addplot +[
      hist={
        bins=11,
        data min=-0.4,
        data max=0.7
      }
    ] table [y index=0] {data/weight_diffs.dat};
  \end{axis}
\end{tikzpicture}
\caption{Histogram of differences between mean pixel weight within
  ($\mu_{\mbox{in}}$) annotated regions and across the whole image ($\mu$). Pixel
  weights are normalized by the maximum pixel weight. More weight is usually assigned to the relevant region: often much more and very rarely much less.
  }
  \label{fig:diff_hist}
\end{figure}

\begin{table}
  \small
  \centering
  \begin{tabular}{l|c|c|c|c}
    & region & image & text & freq\\
    \hline
overall & \textbf{58.94} & 57.83 & 53.98 & 100.0\%\\
is/are/was & \textbf{75.42} & 74.63 & 75.00 & 33.3\%\\
identify: what & \textbf{52.89} & 52.10 & 45.11 & 23.8\%\\
\multicolumn{1}{r|}{kind/type/animal} & &&&\\
how many & 33.38 & \textbf{36.84} & 34.05 & 10.3\%\\
what color & \textbf{53.96} & 43.52 & 32.59 & 9.8\%\\
interpret: & \textbf{75.73} & 74.43 & \textbf{75.73} & 4.6\%\\
\multicolumn{1}{r|}{can/could/does/has} &&&&\\
none of the above  & 45.40 & 44.04 & \textbf{48.23} & 4.1\%\\
where & 42.11 & \textbf{42.43} & 37.61 & 2.5\%\\
why/how & 26.31 & 28.18 & \textbf{29.24} & 2.2\%\\
relational: what is& \textbf{70.15} & 67.48 & 56.64 & 2.0\%\\
\multicolumn{1}{r|}{the man/woman}  &&&&\\
relational: what is & 54.78 & \textbf{54.80} & 45.41 & 1.8\%\\
\multicolumn{1}{r|}{in/on} &&&&\\
which/who & \textbf{43.97} & 42.70 & 38.62 & 1.7\%\\
reading: what & \textbf{33.31} & 31.54 & 30.84 & 1.6\%\\
\multicolumn{1}{r|}{does/number/name}  &&&&\\
identify scene: & \textbf{86.21} & 76.65 & 61.26 & 0.9\%\\
\multicolumn{1}{r|}{what room/sport} &&&&\\
what time  & \textbf{41.47} & 37.74 & 38.64 & 0.8\%\\
what brand  & 45.40 & 44.04 & \textbf{48.23} & 0.4\%\\
  \end{tabular}
   \caption{Accuracies by type of question on the validation set.
     Percent accuracy is shown for each subset for our proposed
     region-based approach, classification using the whole image and
     question/answer text, and classification based only on the text.
     We also show the frequency of each question type.   Note that
     since there are 121,512 questions used for testing, there
     are hundreds or thousands of examples of even the rarest question
     types, so small gains are statistically meaningful.  Overall, our
     region selection scheme outperforms use of whole images by $2\%$
     and text-only features by $5\%$.  There is substantial
     improvement in particular types of questions.  For example,
     questions such as ``What is the woman holding?'' are answered
     correctly 70\% of the time vs. 67\% for whole image and only 57\%
     for text.  ``What color,'' ``What room,'' and ``What sport'' also
     benefit greatly from use of image features and further from
     region weighting.  Question types that have yes/no answers tend
     not to improve, in part because the prior is so reliable.  E.g.,
     someone is unlikely to ask ``Does the girl have a lollipop?'' if
     she is not so endowed.  So ``no'' answers are unlikely and also
     more difficult to verify. We also note that reading questions
     (``What does the sign say?'') and counting questions (``How many
     sheep?'') are not greatly improved by visual features in our system because they require specialized processes.
   }
\label{tbl:breakdown}
\end{table}

\subsection{Region Evaluation}

\begin{table}[ht]
  \centering
\begin{tabular}{l|c}
  Model & Accuracy (\%)\\
  \hline
  Q+A (2-bin) & 51.87 \\
  parsed(Q)+A (5-bin) & \textbf{53.98} \\
\end{tabular}
\caption{Language model comparison. The 2-bin model is the
  concatenation of the question and answer averages. The parsed model
  uses the Stanford dependency parser to further split the question
  into 4 bins.}
\label{tbl:language}
\end{table}

In order to evaluate the consistency of our region weightings with
respect to various types of questions, we set up an informal
experiment to directly evaluate them. To determine how well the region
weighting corresponded to regions a person would use to answer a
question, we manually annotated 205 images from the validation set with bounding
boxes considered important to answer the corresponding
question. An example of the annotation and predicted weights can be
seen in Fig.~\ref{fig:weight_anno}. To evaluate, we compare the average pixel weighting within the
annotated boxes with the average across all pixels. Pixel weighting
was determined by cumulatively adding each region's selection weight to each of its
constituent pixels. We observe that the the mean weighting within the
annotated regions was greater than the global average in 148 of the
instances (72.2\%), often much greater, and rarely much smaller (Fig.~\ref{fig:diff_hist}).

\subsection{Language Model}
We also compare our parsed and binned language model with a simple two-binned
model (one bin averages word2vec of question words; the other averages answer words) to justify our more complex representation. Each
model is trained on the train set and evaluated on the validation set
of the VQA real-images subset. The comparison results are shown in
Table~\ref{tbl:language} and depict a significant performance
improvement using the parsing.

\section{Conclusion}

We presented a model that learns to select regions from the image to solve visual question answering  problems. Our model outperforms all baselines and existing work on the MS COCO VQA multiple choice
task~\cite{VQA}, with substantial gains for some questions such as identifying object colors that require focusing on particular regions.  One direction for future work is to learn to perform specialized tasks such as counting or reading.  Other directions are to incorporate and adapt pre-trained models for object and attribute detectors or geometric reasoning, or to use outside knowledge sources to help learn what is relevant to answer difficult questions.  We are also interested in learning where to look to find small objects and recognize activities.

\clearpage


{\small
\bibliographystyle{ieee}
\bibliography{egbib}
}

\end{document}